\documentclass[10pt,twocolumn,letterpaper]{article}

\usepackage{cvpr}
\usepackage{times}
\usepackage{epsfig}
\usepackage{graphicx}
\usepackage{subfig}
\usepackage{tabularx}
\usepackage{multirow}
\usepackage{float}
\usepackage{amsmath}
\usepackage{amssymb}
\usepackage{booktabs}
\usepackage{paralist}

\newcommand{\specialcell}[2][c]{%
  \begin{tabular}[#1]{@{}c@{}}#2\end{tabular}}

\usepackage[pagebackref=true,breaklinks=true,letterpaper=true,colorlinks,bookmarks=false]{hyperref}

\cvprfinalcopy 

\def\cvprPaperID{2605} 

\ifcvprfinal\fi
\begin{document}


\title{Beyond the Pixel-Wise Loss for Topology-Aware Delineation}

\author{Agata Mosinska\textsuperscript{1}\thanks{This work was supported in part by SNF}
\quad
Pablo M\'{a}rquez-Neila\textsuperscript{1}\textsuperscript{2}
\quad
Mateusz Kozi\'{n}ski\textsuperscript{1}
\quad
Pascal Fua\textsuperscript{1}\\
\textsuperscript{1}Computer Vision Laboratory, \'{E}cole Polytechnique F\'{e}d\'{e}rale de Lausanne (EPFL)\\
\textsuperscript{2}ARTORG Center for Biomedical Engineering Research, University of Bern\\
{\tt\small \{agata.mosinska, pablo.marquezneila, mateusz.kozinski, pascal.fua\}@epfl.ch}
}

\maketitle


\newif\ifdraft
\drafttrue

\definecolor{orange}{rgb}{1,0.5,0}

\ifdraft
 \newcommand{\PF}[1]{{\color{red}{\bf PF: #1}}}
 \newcommand{\pf}[1]{{\color{red} #1}}
 \newcommand{\AM}[1]{{\color{blue}{\bf AM: #1}}}
 \newcommand{\am}[1]{{\color{blue} #1}}
 \newcommand{\PMN}[1]{{\color{orange}{\bf PMN: #1}}}
 \newcommand{\pmn}[1]{{\color{orange} #1}}
  \newcommand{\MK}[1]{{\color{cyan}{\bf MK: #1}}}
 \newcommand{\mk}[1]{{\color{cyan} #1}}

\else
 \newcommand{\PF}[1]{{\color{red}{}}}
 \newcommand{\pf}[1]{ #1 }
 \newcommand{\AM}[1]{{\color{blue}{}}}
 \newcommand{\am}[1]{ #1 }
 \newcommand{\PMN}[1]{{\color{red}{}}}
 \newcommand{\pmn}[1]{ #1 }
\fi

\newcommand{\bW}[0]{\mathbf{W}}
\newcommand{\bw}[0]{\mathbf{w}}
\newcommand{\bL}[0]{\mathcal{L}}
\newcommand{\bX}[0]{\mathbf{X}}
\newcommand{\bx}[0]{\mathbf{x}}
\newcommand{\bY}[0]{\mathbf{Y}}
\newcommand{\by}[0]{\mathbf{y}}
\newcommand{\hby}[0]{\hat{\mathbf{y}}}
\newcommand{\iw}[0]{\mathit{w}}

\newcommand{\real}{\mathbb{R}}

\newcommand{\MNIH}[0]{\textbf{MNIH}}
\newcommand{\LDNN}[0]{\textbf{CHM-LDNN}}
\newcommand{\UNET}[0]{\textbf{U-Net}}
\newcommand{\REGC}[0]{\textbf{Reg-AC}}
\newcommand{\CrackTree}[0]{\textbf{CrackTree}}

\newcommand{\EMdat}[0]{\textbf{EM}}
\newcommand{\RDdat}[0]{\textbf{Roads}}
\newcommand{\CRdat}[0]{\textbf{Cracks}}

\newcommand{\OURS}[0]{\textbf{OURS}}
\newcommand{\OURN}[0]{\textbf{OURS-NoRef}}

\newcommand{\comment}[1]{}


\begin{abstract}

Delineation of curvilinear structures is an important problem in Computer Vision with multiple practical applications. With the advent of Deep Learning, many current approaches on automatic delineation have focused on finding more powerful deep architectures, but have continued using the habitual pixel-wise losses such as binary cross-entropy. In this paper we claim that pixel-wise losses alone are unsuitable for this problem because of their inability to reflect the topological impact of mistakes in the final prediction. We propose a new loss term that is aware of the higher-order topological features of linear structures. We also introduce a refinement pipeline that iteratively applies the same model over the previous delineation to refine the predictions at each step while keeping the number of parameters and the complexity of the model constant.

When combined with the standard pixel-wise loss, both our new loss term and our iterative refinement boost the quality of the predicted delineations, in some cases almost doubling the accuracy as compared to the same classifier trained with the binary cross-entropy alone. We show that our approach outperforms state-of-the-art methods on a wide range of data, from microscopy to aerial images.

\end{abstract}


\section{Introduction}

\begin{figure}
  \centering
  \begin{tabular}{cc}
\hspace{-3mm}\includegraphics[width=0.23\textwidth]{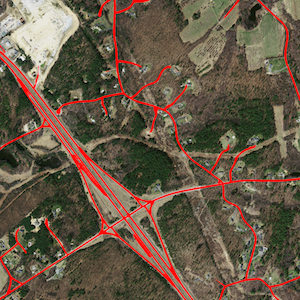}&
\hspace{-3mm}\includegraphics[width=0.23\textwidth]{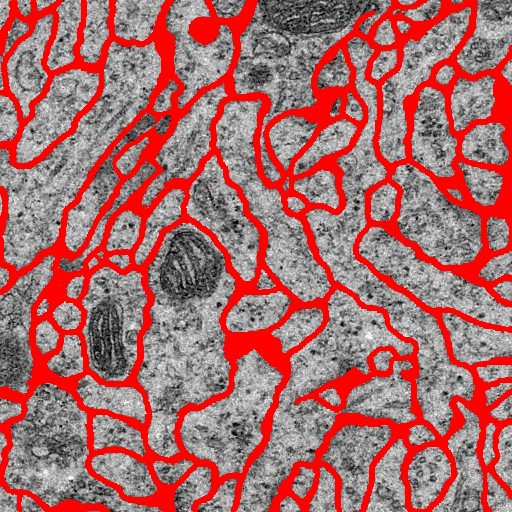} \\
(a) & (b) \\
 \includegraphics[width=0.23\textwidth]{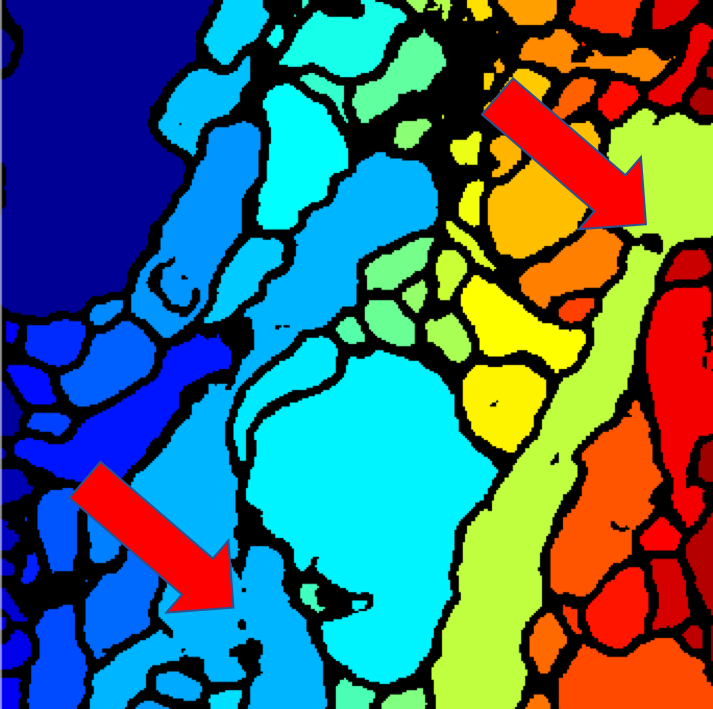} &
 \includegraphics[width=0.23\textwidth]{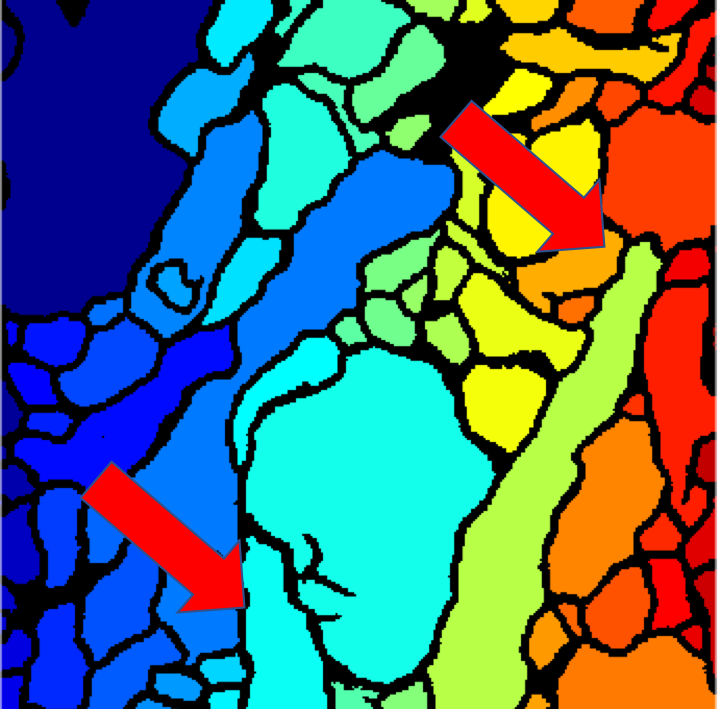} \\
 (c) & (d)
 \end{tabular}
  \vspace{-3mm}
  \caption{\textbf{Linear structures.} (a) Detected roads in an aerial image. (b) Detected cell membranes in an electron microscopy (EM) image. (c)  Segmentation obtained after detecting neuronal membranes using~\cite{Ronneberger15} (d) Segmentation obtained after detecting membranes using our method. Our approach closes small gaps, which prevents much bigger topology mistakes.}
  \label{fig:teaser}
\end{figure}

Automated delineation of curvilinear structures, such as those in Fig.~\ref{fig:teaser}(a, b), has been investigated since the inception of the field of Computer Vision in the 1960s and 1970s. Nevertheless, despite decades of sustained effort, full automation remains elusive when the image data is noisy and the structures are complex. As in many other fields, the advent of Machine Learning techniques in general, and Deep Learning in particular, has produced substantial advances, in large part because learning features from the data makes them more robust to appearance variations~\cite{Ganin2014,Mnih12,Sironi16a,Wegner13}.

However, all new methods focus on finding either better features to feed a classifier or more powerful deep architectures while still using a pixel-wise loss such as  binary cross-entropy for training purposes. Such loss is entirely local and does not account for the very specific and sometimes complex topology of curvilinear structures penalizing all mistakes equally regardless of their influence on geometry.  As a shown in Fig.~\ref{fig:teaser}(c,d) this is a major problem because small localized pixel-wise mistakes can result in large topological changes.

In this paper, we show that supplementing the usual pixel-wise loss by a {\it topology loss} that promotes results with appropriate topological characteristics makes a big difference and yields a substantial performance increase without having to change the network architecture. In practice, we exploit the feature maps computed by a pretrained VGG19~\cite{Simonyan15} to obtain high-level descriptions that are sensitive to linear structures. We use them to compare the topological properties of the ground truth and the network predictions and estimate our topology loss.

In addition to this, we introduce an iterative refinement pipeline inspired by recent detection and segmentation methods~\cite{Newell16,Shen17}. We show that, unlike in these earlier methods, we can share the architecture and the parameters along all the refinement steps instead of instantiating a new network each time.  This keeps the number of parameters constant irrespective of the number of iterations, which is important when only a relatively small amount of training data is available, as is often the case in biomedical and other specialized applications.
In our experiments, we will show that our topology loss, together with the iterative refinement approach, can boost the accuracy of a deep network by up to~30~percent points as compared to the same network trained in the same way but using a standard pixel-wise loss and without iterative refinement.

Our main contribution is therefore a demonstration that properly accounting for topology in the loss used to train the network is an important step in boosting performance.


\section{Related Work}
\label{sec:related}

\subsection{Detecting Linear Structures}

Delineation algorithms can rely either on hand-crafted or on learned features. Optimally Oriented Flux (OOF)~\cite{Law08} and Multi-Dimensional Oriented Flux (MDOF)~\cite{Turetken13c}, its extension to irregular structures,  are successful examples of the former. Their great strength is that they do not require training data but at the cost of struggling  with very irregular structures at different scales along with the great variability of appearances and artifacts.

In such challenging situations, learning-based methods have an edge and several approaches have been proposed over the years. For example, the approach of~\cite{Zhou12} combines Haar wavelets with boosted trees, while that of~\cite{Huang09} performs SVM binary classification on spectral features. In~\cite{Sironi16a}, the classifier is replaced by a regressor that predicts the distance to the closest centerline, which enables estimating the width of the structures. 

In more recent work, Decision Trees and SVMs have been replaced by Deep Networks.  For the purpose of road delineation, this was first done in~\cite{Mnih10}, directly using image patches as input to a  fully connected neural net. While the patch provided some context around the linear structures, it was still relatively small due to memory limitations. With the advent of Convolutional Neural Networks (CNNs), it became possible to use larger receptive fields. In~\cite{Ganin2014}, CNNs were used to extract features that could then be matched against \emph{words} in a learned dictionary. The final prediction was made based on the votes from nearest neighbors in the feature space. A fully-connected network was replaced by a CNN in~\cite{Mnih13} for road detection. In~\cite{Mattyus17} a differentiable Intersection-over-Union loss was introduced to obtain a road segmentation, which is then used to extract graph of the road network.

In the biomedical field, the VGG network~\cite{Simonyan15}  pre-trained on real images has been fine-tuned and augmented by specialized layers to extract blood vessels~\cite{Maninis16}. Similarly the U-Net~\cite{ Ronneberger15}, has been shown to give excellent results for biomedical image segmentation and is currently among the methods that yield the best results for neuron boundaries detection in the ISBI'12 challenge~\cite{arganda15}. It is made of an encoder, that is, a contracting path that captures context, followed by an expanding decoder, which localizes objects. It comprises a series of convolutional layers with an either increasing or decreasing number of channels, interleaved with pooling operations for the encoding layers and up-convolutions for the decoding layers. Skip connections between corresponding layers in analysis and synthesis paths provide high-resolution features to the latter.

While effective, all these approaches rely on a standard cross entropy loss for training purposes. Since they operate on individual pixels as though they were independent of each other, they ignore higher-level statistics while scoring the output. We will see in Section~\ref{sec:results} that this is detrimental even when using an architecture designed to produce a structured output, such as the U-Net. 

Of course, topological knowledge can be imposed in the output of these linear structure detectors. For example, in~\cite{Wegner13}, this is done by introducing a CRF formulation whose priors are computed on higher-order cliques of connected superpixels likely to be part of road-like structures. Unfortunately, due to the huge number of potential cliques, it requires sampling and hand-designed features. Another approach to model higher-level statistics is to represent linear structures as a sequence of short linear segments, which can be accomplished using a Marked Point Process~\cite{Stoica04}~\cite{Chai13}. However, the inference involves Reversible Jump Markov Chain Monte Carlo, which is computationally expensive and relies on a very complex objective function. More recently, it has been shown that the delineation problem could be formulated in terms of finding an optimal subgraph in a graph of potential linear structures by solving an Integer Program~\cite{Turetken16a} . However, this requires a complex pipeline whose first step is finding points on the centerline of candidate linear structures. 

\subsection{Recursive Refinement}

Recursive refinement of a segmentation has been extensively investigated. It is usually implemented as a procedure of iterative predictions~\cite{Mnih10,Pinheiro14,Tu09}, sometimes at different resolutions~\cite{Seyedhosseini13}. Such methods use the prediction from a previous iteration (and sometimes the image itself) as the input to a classifier that produces the next prediction. This enables the classifier to better consider the context surrounding a pixel when trying to assign a label to it and has been successfully used for delineation purposes~\cite{Sironi16a}.

In more recent works, the preferred approach to refinement with Deep Learning is to stack several deep modules and train them in an end-to-end fashion. For example, the pose estimation network of~\cite{Newell16} is made of eight consecutive \emph{hourglass} modules and supervision is applied on the output of each one during training, which takes several days.  In~\cite{Shen17} a similar idea is used to detect neuronal membranes in electron microscopy images, but due to memory size constraints the network is limited to 3~modules. In other words, even though such end-to-end solutions are convenient, the growing number of network parameters they require can become an issue when time, memory, and available amounts of training data are limited. This problem is tackled in~\cite{Januszewski16} by using a single network that moves its attention field within the volume to be segmented. It predicts the output for the current field of view and fills in the prediction map. Both the image and the updated prediction map are used as the input to the network in the next iteration. The network weights are then updated after each new prediction and the output prediction is thresholded before serving as an input for the next iteration in order to avoid saturating the network. The method also requires pre-computed seeds to initialize the prediction map at the first iteration. As will be discussed in Section~\ref{sec:refinement},  our approach to refinement also uses a single network and recursively refines its output.  However, in training, we use a loss function that is a weighted sum of losses computed after each processing step. This enables us to accumulate the gradients and requires neither seeds for initialization nor processing the intermediate output.


\section{Method}
\label{sec:method}


\begin{figure*}[t]
  \centering
\begin{tabular}{ccccc}
  \includegraphics[width=0.17\textwidth]{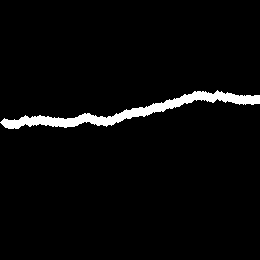} &
  \includegraphics[width=0.17\textwidth]{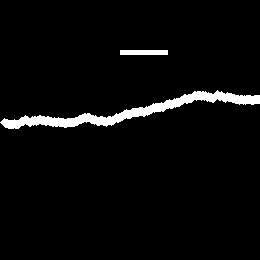} &
  \includegraphics[width=0.17\textwidth]{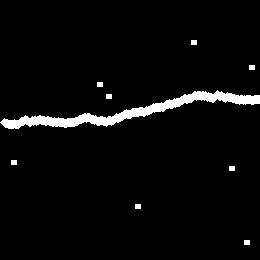} &
  \includegraphics[width=0.17\textwidth]{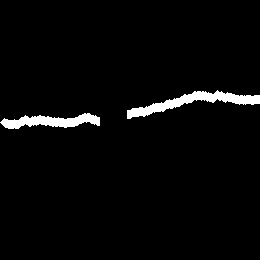} &
   \includegraphics[width=0.17\textwidth]{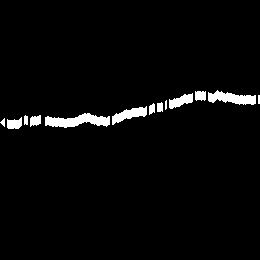}\\
 ground truth & $\bL_{top} = 0.2279$ & $\bL_{top} = 0.7795$ & $\bL_{top} = 0.2858$ & $\bL_{top} = 0.9977$\\
    (a)&(b)&(c)&(d)&(e)\\
\end{tabular}
  \caption{{\bf The effect of mistakes on topology loss.} {\small (a) Ground truth (b)-(e) we flip 240 pixels in each prediction, so that $\bL_{bce}$ is the same for all of them, but as we see $\bL_{top}$ penalizes more the cases with more small mistakes, which considerably change the structure of the prediction.}}
  \label{fig:topo_loss}
\end{figure*}

\begin{figure}
  \centering
\begin{tabular}{ccc}
  \hspace{-5mm} \includegraphics[width=0.16\textwidth]{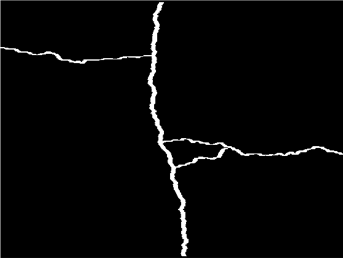} &
  \hspace{-4mm} \includegraphics[width=0.16\textwidth]{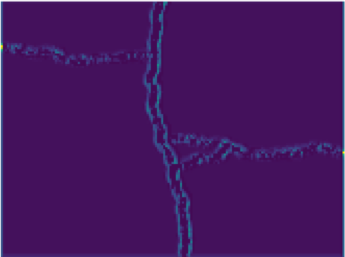} &
  \hspace{-4mm} \includegraphics[width=0.16\textwidth]{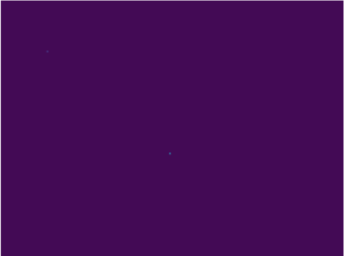} \\
  \hspace{-5mm} \includegraphics[width=0.16\textwidth]{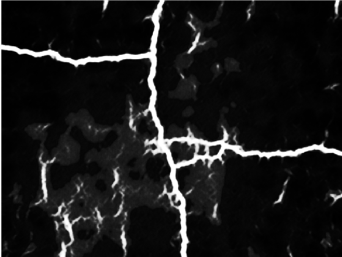} &
  \hspace{-4mm} \includegraphics[width=0.16\textwidth]{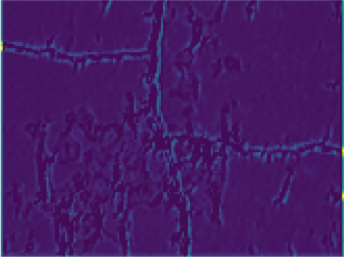} &
  \hspace{-4mm} \includegraphics[width=0.16\textwidth]{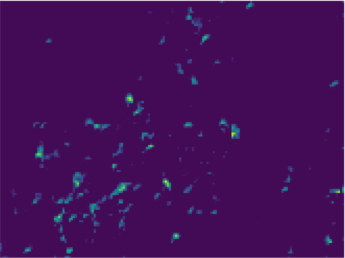} \\
  \hspace{-5mm} (a) & (b) & (c) \\
\end{tabular}
\vspace{-3mm}
  \caption{{\bf Examples of activations in VGG layers.} {\small (a) Ground-truth and corresponding prediction with errors. (b) Responses of a VGG19 channel specialized in elongated structures. (c) Responses of a VGG19 channel specialized in small connected components. $\bL_{top}$~strongly encourages responses in the former and penalizes responses in the latter.}}
  \label{fig:activations}
\end{figure}


We use the fully convolutional U-Net~\cite{Ronneberger15} as our trainable model, as it is currently among the best and most widely used architectures for delineation and segmentation in both natural and biomedical images.
The U-Net is usually trained to predict the probability of each pixel of being a linear structure using a standard pixel-wise loss. As we have already pointed out, this loss relies on local measures and does not account for the overall geometry of curvilinear structures, which is what we want to remedy.

In the remainder of this section, we first describe our topology-aware loss function, and we then introduce our iterated procedure to recursively refine our predictions.

\subsection{Notation}

In the following discussion, let $\bx \in \real^{H\cdot{}W}$~be the $W\times{}H$ input image, and let $\by \in\{0, 1\}^{H\cdot{}W}$~be the corresponding ground-truth labeling, with $1$~indicating pixels in the curvilinear structure and $0$~indicating background pixels.

Let $f$~be our U-Net parameterized by weights~$\bw$. The output of the network is an image~$\hby = f(\bx, \bw) \in [0, 1]^{H\cdot{}W}$.\footnote{For simplicity and without loss of generality, we assume that $\bx$ and~$\hby$ have the same size. This is not the case in practice, and usually $\hby$ corresponds to the predictions of a cropped area of $\bx$ (see~\cite{Ronneberger15} for details).} Every element of $\hby$ is interpreted as the  probability of pixel~$i$ having label~$1$: ~$\hby_i \equiv p(Y_i=1 \mid \bx, \bw)$, where $Y_i$ is a random Bernoulli variable~$Y_i \sim \textrm{Ber}(\hby_i)$.

\subsection{Topology-aware loss}
\label{sec:topologyLoss}

In ordinary image segmentation problems, the loss function used to train the network is usually the standard pixel-wise binary cross-entropy (BCE):
\begin{align}
\nonumber
\bL_{bce}(\bx, \by, \bw) = -\sum_{i} & \left[(1-\by_i)\cdot\log (1-f_i(\bx, \bw))\right. \\
     &\left.+ \by_i\cdot\log f_i(\bx, \bw)\right].
\label{eq:bceLoss}
\end{align}
Even though the U-Net computes a structured output and considers large neighborhoods, this loss function treats every pixel independently. It does not capture the characteristics of the topology, such as the number of connected components or number of holes. This is especially important in the delineation of thin structures: as we have seen in Fig.~\ref{fig:teaser}(c, d), the misclassification of a few pixels might have a low cost in terms of the pixel-wise BCE loss, but have a large impact in the topology of the predicted results.

Therefore, we aim to introduce a penalty term in our loss function to account for this higher-order information. Instead of relying on a hand-designed metric, which is difficult to model and hard to generalize to different image modalities, we leverage the knowledge that a pretrained network contains about the structures of real-world images. In particular, we use the feature maps at several layers of a VGG19 network~\cite{Simonyan15} pretrained on the ImageNet dataset~\cite{Russakovsky15} as a description of the higher-level features of the delineations. Our new penalty term tries to minimize the differences between the VGG19 descriptors of the ground-truth images and the corresponding predicted delineations:
\begin{equation}
\bL_{top}(\bx, \by, \bw) = \sum_{n=1}^N \sum_{m=1}^{M_n} \left\| l^{m}_n (\by) - l^{m}_n (f(\bx, \bw)) \right\|^2_2 \; ,
\label{eq:topoLoss}
\end{equation}
where $l^{m}_n$ denotes the $m$-th feature map in the $n$-th layer of the pretrained VGG19 network, $N$~is the number of convolutional layers considered and $M_n$ is the number of channels in the $n$-th feature map. $\bL_{top}$ can be understood as a measurement of the difference between the higher-level visual features of the linear structures in the ground-truth and those in predicted image. These higher-level features include concepts such as connectivity or holes that are ignored by the simpler pixel-wise BCE loss. Fig.~\ref{fig:topo_loss} shows examples where the pixel-wise loss is too weak to penalize properly a variety of errors that occur in the predicted delineations, while our loss $\bL_{top}$~correctly measures the topological importance of the errors in all cases: it penalizes more the mistakes that considerably change the structure of the image and those that do not resemble linear structures.

The reason behind the good performance of the VGG19 in this task can be seen in Fig.~\ref{fig:activations}. Certain channels of the VGG19 layers are activated by the type of elongated structures we are interested in, while others respond strongly to small connected components. Thus, minimizing~$\bL_{top}$ strongly penalizes generating small false positives, which do not exist in the ground-truth, and promotes the generation of elongated structures. On the other hand, the shape of the predictions is ignored by~$\bL_{bce}$.

An interesting direction for future research would be learning $\bL_{top}$ instead of defining it with a pretrained model. However, this bears some resemblance to Generative Adversarial Networks (GAN)~\cite{Goodfellow14}, whose training is known to be prone to problems such as \emph{mode collapse} and instability. They are also known to be hard to train for segmentation purposes~\cite{Isola17}. 

In the end, we minimize
\begin{equation}
\bL (\bx, \by, \bw) = \bL_{bce}(\bx, \by, \bw) + \mu\bL_{top}(\bx, \by, \bw)
\label{eq:newloss}
\end{equation}
with respect to $\bw$. $\mu$~is a scalar weighing the relative influence of both terms. We set it so that the order of magnitude of both terms is comparable. Fig.~\ref{fig:net}(a) illustrates the proposed approach.


\begin{figure*}[t]
  \centering
  \begin{tabular}{cc}
  \includegraphics[width=0.4\textwidth]{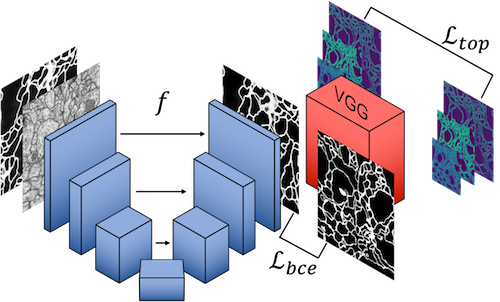}\label{fig:net}&
  \includegraphics[width=0.5\textwidth]{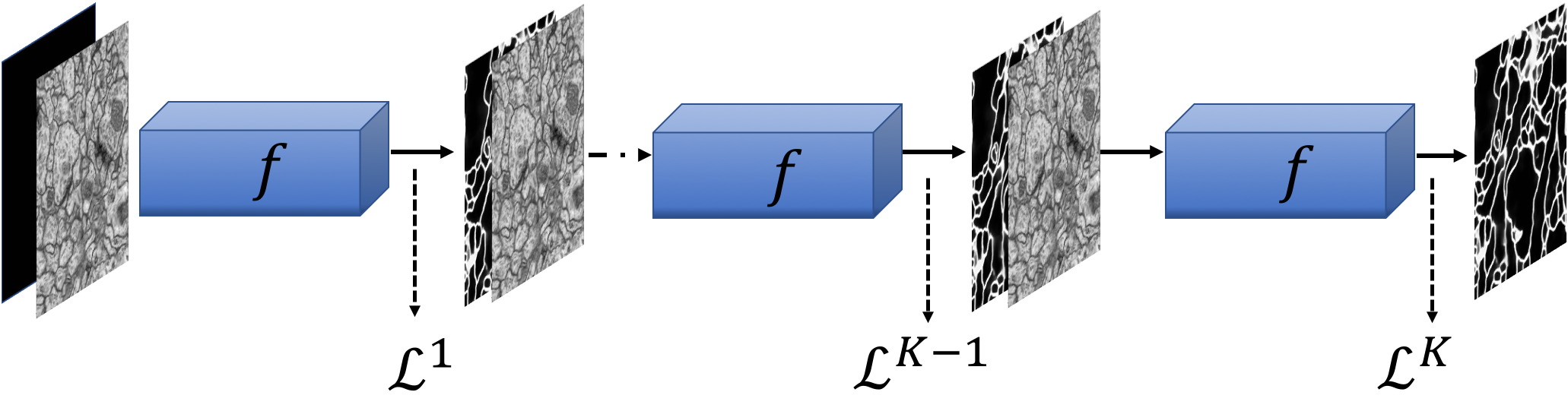}\label{fig:refinement}\\
  (a)&(b)
  \end{tabular}
  \vspace{-1mm}
  \caption{{\bf Network architecture.} {\small (a) We use the U-Net for delineation purposes. During training, both its output and the ground-truth image serve as input to a pretrained VGG network.
  The loss $\bL_{top}$ is computed from the VGG responses. The loss $\bL_{bce}$ is computed pixel-wise between the prediction and the ground-truth. (b) Our model iteratively applies the same U-Net~$f$ to produce progressive refinements of the predicted delineation. The final loss is a weighted sum of partial losses~$\bL^k$ computed at the end of each step.}}
  \label{fig:net}
\end{figure*}

\subsection{Iterative refinement}
\label{sec:refinement}

The topology loss term of Eq.~\ref{eq:topoLoss} improves the quality of the predictions. However, as we will see in Section~\ref{sec:results}, some mistakes still remain. They typically show up in the form of small gaps in lines that should be uninterrupted. We iteratively refine the predictions to eliminate such problems. At each iteration, the network takes both the input image and the prediction of the previous iteration to successively provide better predictions. In earlier works that advocate a similarly iterative approach~\cite{Newell16,Shen17}, a different module~$f^k$ is trained for each iteration~$k$, thus increasing the number of parameters of the model and making training more demanding in terms of the amount of required labeled data. An interesting property of this iterative approach is that the correct delineation $\by$~should be the fixed point of each module~$f^k$, that is, feeding the correct delineation should return the input
\begin{equation}
\by = f^k(\bx \oplus \by),
\end{equation}
where $\oplus$~denotes channel concatenation and we omitted the weights of~$f^k$ for simplicity. Assuming that every module~$f^k$ is Lipschitz-continuous on~$\by$,\footnote{Lipschitz continuity is a direct consequence of the assumption that every~$f^k$ will always improve the prediction of the previous iteration.} we know that the fixed-point iteration
\begin{equation}
f^k(\bx \oplus f^k(\bx \oplus f^k(\ldots)))
\end{equation}
converges to~$\by$.
We leverage this fixed-point property to remove the necessity of training a different module at each iteration. Instead, we use the same single network~$f$ at each step of the refinement pipeline, as depicted in Fig.~\ref{fig:net}(b). This makes our model much simpler and less demanding of labeled data for training. Also, it helps the network to learn a contraction map that successively improves the estimations.
Our predictive model can therefore be expressed as
\begin{equation}
\hby^{k+1} = f(\bx \oplus \hby^k, \bw), \quad k=0,\ldots,K-1 \; ,
\label{eq:refine}
\end{equation}
where $K$ is the total number of iterations and $\hby^K$ the final prediction. We initialize the model with an empty prediction $\hby^0 = \mathbf{0}$.

Instead of minimizing only the loss for the final network output, we minimize a weighted sum of partial losses. The $k$-th partial model, with $k\le K$, is the model obtained from iterating Eq.~\ref{eq:refine} $k$~times. The $k$-th partial loss~$\bL^k$ is the loss from Eq.~\ref{eq:newloss} evaluated for the $k$-th partial model. Using this notation, we define our {\it refinement} loss as a weighted sum of the partial losses

\begin{equation}
\bL_{ref}(\bx, \by, \bw) = \dfrac{1}{Z} \sum_{k=1}^{K} k\,\bL^k(\bx, \by, \bw) \; ,
\label{eq:lossRef}
\end{equation}
with the normalization factor $Z = \sum_{k=1}^{K}k=\frac{1}{2}K(K+1)$.
We weigh more the losses associated with the final iterations to boost the accuracy of the final result. However, accounting for the earlier losses enables the network to learn from all the mistakes it can make along the way and increases numerical stability.  It also avoids having to preprocess the predictions before re-injecting them into the computation, as in~\cite{Januszewski16}.  

In practice, we first train a single  module network, that is, for $K=1$.  We then increment $K$, retrain, and iterate. We limit $K$ to 3 during training and testing as the results do not change significantly for larger $K$~values. We will show that this successfully fills in small gaps while removing background noise.


\section{Results}
\label{sec:results}

\begin{figure*}
  \centering
  \includegraphics[width=0.85\textwidth]{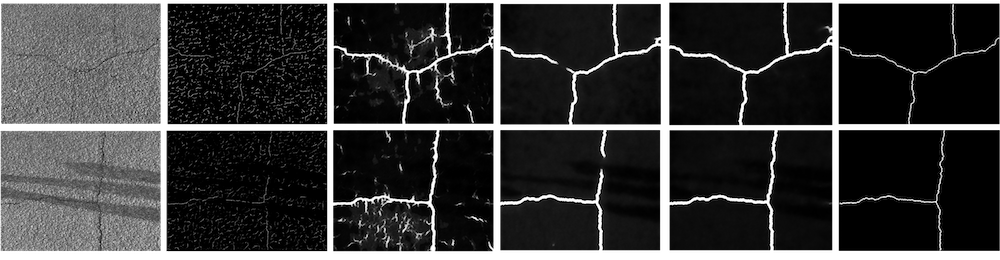}
   \vspace{-2mm}
  \caption{\small \textbf{Cracks.} From left to right: image, \REGC, \UNET , \OURN{} and \OURS{} prediction, ground-truth.}
  \label{fig:cracks_qualitative}
\end{figure*}

\begin{figure*}
  \centering
  \includegraphics[width=0.85\textwidth]{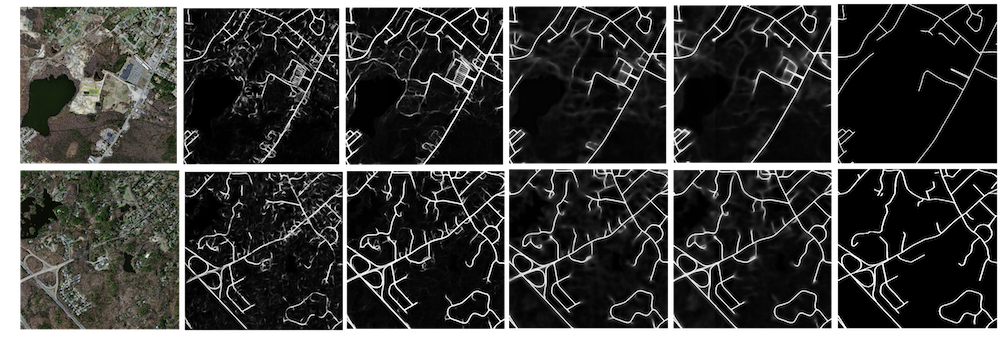}
   \vspace{-2mm}
  \caption{\small \textbf{Roads}. From left to right: image, \MNIH, \UNET , \OURN{} and \OURS{} prediction, ground-truth.}
  \label{fig:roads_qualitative}
\end{figure*}

\begin{figure*}
  \centering
  \includegraphics[width=0.85\textwidth]{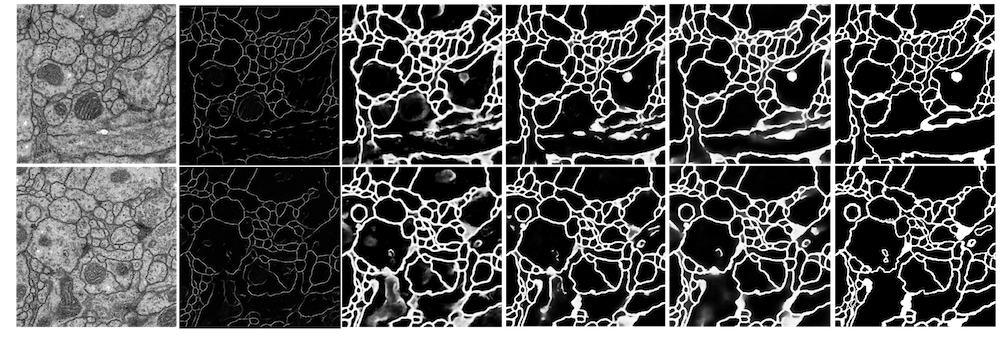}
   \vspace{-2mm}
  \caption{\small \textbf{EM}.  From left to right: image, \REGC, \UNET , \OURN{} and \OURS{} prediction, ground-truth.}
  \label{fig:isbi_qualitative}
\end{figure*}

\paragraph{Data.}
We evaluate our approach against state-of-the-art methods~\cite{Mnih13,Ronneberger15,Seyedhosseini13,Sironi16a} on three datasets featuring very different kinds of linear structures:
\begin{compactenum}

\item \textbf{Cracks:} Images of cracks in asphalt~\cite{Zou12}. It consists of 104 training and 20 test images. As can be seen in Fig.~\ref{fig:cracks_qualitative}, the multiple shadows and cluttered background makes their detection a challenging task. Potential applications include quality inspection and material characterization.

\item \textbf{Roads:} The Massachusetts Roads Dataset~\cite{Mnih13} is one of the largest publicly available collections of aerial road images, containing both urban and rural neighbourhoods, with many different kinds of roads ranging from small paths to highways.
The set is split into 1108 training and 49 test images, 2 of which are shown in Fig.~\ref{fig:roads_qualitative}. 

\item \textbf{EM:} We detect neuronal boundaries in Electron Microscopy images from the ISBI'12 challenge~\cite{arganda15} (Fig.~\ref{fig:isbi_qualitative}). There are 30 training images, with ground truth annotations, and 30 test images for which the ground-truth is withheld by the organizers. Following~\cite{Seyedhosseini13}, we split the training set into 15 training and 15 test images. We report our results on this split.

\end{compactenum}
\paragraph{Training protocol.}
Since the U-Net cannot handle very large images, we work with patches of $450 \times 450$~pixels for training. We perform data augmentation mirroring and rotating the training images by 90$^{\circ}$, 180$^{\circ}$ and 270$^{\circ}$. Additionally, in the \textbf{EM} dataset, we also apply elastic deformations as suggested in~\cite{Ronneberger15} to compensate for the small amount of training data. We use batch normalization for faster convergence and use current batch statistics also at the test time as suggested in~\cite{Cicek16}. We chose Adam~\cite{Kingma15} with a learning rate of~$10^{-4}$ as our optimization method. 

\label{sec:metrics}

\vspace{-3mm}
\paragraph{Pixel-wise metrics.}

Our algorithm outputs a probabilty map, which lends itself to evaluation in terms of precision- and recall-based metrics, such as the F1 score~\cite{Seyedhosseini13} and the precision-recall break-even point~\cite{Mnih13}. 
They are well suited for benchmarking binary segmentations, but their local character is a drawback in the presence of thin structures. Shifting a  prediction even by a small distance in a direction perpendicular to the structure yields zero precision and recall, while still reasonably representing the data. We therefore evaluate the results in terms of \emph{correctness}, \emph{completeness}, and \emph{quality}~\cite{Wiedemann98}.  They are measures designed  specifically for linear structures. They measure the similarity between predicted skeletons and ground truth-ones. They are more sensitive to alignments of centerlines than to precise locations or small width changes of the underlying structures. Potential shifts in centerline positions are handled by relaxing the notion of a true positive from being a  precise coincidence of points to not exceeding a distance threshold. Correctness corresponds to relaxed precision, completeness to relaxed recall, and quality to intersection-over-union. We give precise definitions in appendix. In our experiments we use a threshold of $2$ pixels for roads and cracks, and $1$ for the neuronal membranes.

\vspace{-3mm}
\paragraph{Topology-based metrics.}

The pixel-wise metrics are oblivious of topological differences between the predicted and ground-truth networks. A more topology-oriented set of measures was proposed in~\cite{Wegner13}. It involves finding the shortest path between two randomly picked connected points in the predicted network and the equivalent path in the ground-truth network. If no equivalent ground truth path exists, the former is classified as \emph{infeasible}. It is classified as \emph{too-long/-short} if the length of the paths differ by more than 10\%, and as \emph{correct} otherwise. In practice, we sample 200 paths per image, which is enough for the proportion of correct, infeasible, and too-long/-short paths to stabilize.

The organizers of the EM challenge use a performance metric called {\it foreground-restricted random score}, oriented at evaluating the preservation of separation between different cells. It measures the probability that two pixels belonging to the same cell in reality also do so in the predicted output. As shown in Fig.~\ref{fig:teaser}, this kind of metric is far more sensitive to topological than to pixel-wise perturbations.

\paragraph{Baselines and variants of the proposed method.}

We compare the results of our method to the following baselines:
\begin{compactitem}
\item \CrackTree~\cite{Zou12} a crack detection method based on segmentation and subsequent graph construction
\item \MNIH~\cite{Mnih13}, a neural network for road segmentation in $64\times 64$ image patches, 
\item \UNET~\cite{Ronneberger15}, pixel labeling using the U-Net architecture with BCE loss,
\item \LDNN~\cite{Seyedhosseini13}, a multi-resolution recursive approach to delineating neuronal boundaries,
\item \REGC~\cite{Sironi16a}, a regression-based approach to finding centerlines and refining the results using autocontext. 
\end{compactitem}
We reproduce the results for \MNIH{}, \UNET{} and \REGC{}, and report the results published in the original work for \LDNN.
We also perform an ablation study to isolate the individual contribution of the two main components of our approach. To this end, we compare two variants of it.
\begin{compactitem}
\item \OURN{}, our approach with the topological loss of Eq.~\ref{eq:newloss} but no refinement steps, that is, $K=1$. To extract global features we use the channels from the VGG layers \textbf{conv1\_2}, \textbf{conv2\_2} and \textbf{conv3\_4}, and set $\mu$ to $0.1$ in Eq.~\ref{eq:newloss}.
\item \OURS{}, our complete, iterative method including the topological term and $K=3$~refinement steps. It is trained using the refinement loss of Eq.~\ref{eq:lossRef} as explained in Section~\ref{sec:refinement}.
\end{compactitem}

\subsection{Quantitative Results}
\label{sec:quantitative}å


\begin{table}
\begin{center}
{\small
\begin{tabular}{l r | l r}
VGG layers & Quality & Number of iterations & Quality \\
\hline
\textbf{None} & 0.4050 & \OURN{} & 0.5580  \\
\textbf{layer 1} & 0.6408  & \OURS{} 1 iteration & 0.5621\\
\textbf{layer 2} & 0.6427 & \OURS{}  2 iterations & 0.5709\\
\textbf{layer 3} & 0.6974 & \OURS{}  3 iterations  & 0.5722  \\
\textbf{layers 1,2,3} & 0.7151 & \OURS{} 4 iterations & 0.5727\\
\end{tabular}
}
 \vspace{-3mm}
 \caption{\small Testing different configurations. (Left) Quality scores for \OURN{} method when using different VGG layers to compute the topology loss of Eq.~\ref{eq:topoLoss} on the \CRdat{} dataset. (Right)  Quality scores for \OURS{} method on the \textbf{EM} dataset as a function of the number of refinement iterations. \OURN{} included for comparison.}
\label{table:config}
\end{center}
 \vspace{-3mm}
\end{table}

\begin{table}
\begin{center}
{\small
  \begin{tabular}{l l | l r}
   Method & P/R & Method & F1 \\
  \hline
  \MNIH~\cite{Mnih13} & 0.6822 & \LDNN{}~\cite{Seyedhosseini13} & 0.8072\\
  \UNET{}~\cite{Ronneberger15} & 0.7460 & \UNET{}~\cite{Ronneberger15} & 0.7952\\
  \OURN{} & 0.7610 & \OURN{} & 0.8140\\
  \OURS & \textbf{0.7782} &  \OURS{}  & \textbf{0.8230}\\
  \end{tabular}
}
\end{center}
\vspace{-5mm}
\caption{\small Experimental results on the \RDdat{} and \EMdat{} datasets. (Left) Precision-recall break-even point (P/R) for the  \RDdat{} dataset. Note the results are expressed in terms of the standard precision and recall, as opposed to the relaxed measures reported in \cite{Mnih13}. (Right) F1 scores for the \EMdat{} dataset.}
\label{table:roadsEM}
\end{table}

\begin{table}
\begin{center}
{\small
\begin{tabular}{l l c c r}
Dataset &  Method & Correct. & Complet.  & Quality\\
\midrule
\multirow{5}{*}{\textbf{Cracks}} & \CrackTree~\cite{Zou12} & 0.7900 & 0.9200 & 0.7392 \\
										& \REGC{}~\cite{Sironi16a}& 0.1070 & 0.9283 & 0.1061\\ 
									   & \UNET{}~\cite{Ronneberger15} & 0.4114 & 0.8936 & 0.3924\\
									   & \OURN{}  & 0.7955 & 0.9208 & 0.7446\\
									   & \OURS{} & \textbf{0.8844} & \textbf{0.9513} & \textbf{0.8461}\\
\cmidrule{2-5}
\multirow{5}{*}{\textbf{Roads}} & \REGC{}~\cite{Sironi16a}& 0.2537 & 0.3478 & 0.1719\\
									   & \MNIH~\cite{Mnih13} & 0.5314 & 0.7517 & 0.4521\\
									   & \UNET{}~\cite{Ronneberger15} & 0.6227 & 0.7506& 0.5152\\
									   & \OURN{} & 0.6782 & 0.7986 & 0.5719\\
									   & \OURS{} & \textbf{0.7743} & \textbf{0.8057} & \textbf{0.6524}\\
\cmidrule{2-5}
\multirow{5}{*}{\textbf{EM}} & \REGC{}~\cite{Sironi16a}& 0.7110 & 0.6647 & 0.5233\\
									   & \UNET{}~\cite{Ronneberger15} & 0.6911 & 0.7128 & 0.5406 \\
									   & \OURN{} & 0.7096 & 0.7231 & 0.5580\\
									   & \OURS{} & \textbf{0.7227} & \textbf{0.7358} & \textbf{0.5722}\\
\end{tabular}
}
\end{center}
 \vspace{-5mm}
\caption{\small Correctness, completeness and quality scores for extracted centerlines.}
\label{table:correctness}
 \vspace{-5mm}
\end{table}

\begin{table}
\begin{center}
{\small
\begin{tabular}{l l c c r}
Dataset &  Method & Correct & Infeasible  & \specialcell{2Long \\ 2Short}\\
\midrule
\multirow{5}{*}{\textbf{Cracks}} & \REGC~\cite{Sironi16a}& 39.7 & 56.8 & 3.5\\
									   & \UNET{}~\cite{Ronneberger15} & 68.4 & 27.4 & 4.2\\
									   & \OURN{}  & 90.8 & 6.1 & 3.1\\
									   & \OURS{} & \textbf{94.3} & 3.1 & 2.6\\
\cmidrule{2-5}
\multirow{5}{*}{\textbf{Roads}} & \REGC~\cite{Sironi16a}& 16.2 & 72.1 & 11.7\\
									   & \MNIH~\cite{Mnih13} & 45.5 & 49.73 & 4.77 \\ 
									   & \UNET{}~\cite{Ronneberger15} & 56.3 & 38.0 & 5.7\\
									   & \OURN{} & 63.4 & 32.3 & 4.3\\
									   & \OURS{} & \textbf{69.1} & 24.2 & 6.7\\
\cmidrule{2-5}
\multirow{5}{*}{\textbf{EM}} & \REGC~\cite{Sironi16a}& 36.1 & 38.2 & 25.7\\
									   & \UNET{}~\cite{Ronneberger15} & 51.5 & 16.0 & 32.5 \\
									   & \OURN{} & 63.2 & 16.8 & 20.0\\
									   & \OURS{} & \textbf{67.0} & 15.5 &17.5\\
\end{tabular}
}
\end{center}
 \vspace{-5mm}
 \caption{\small The percentage of correct, infeasible and too-long/too-short paths sampled from predictions and ground truth.}
\label{table:topology}
 \label{table:topology}
 \vspace{-5mm}
\end{table}


\begin{figure*}
  \centering
  \begin{tabular}{c}
  \includegraphics[width=0.60\textwidth]{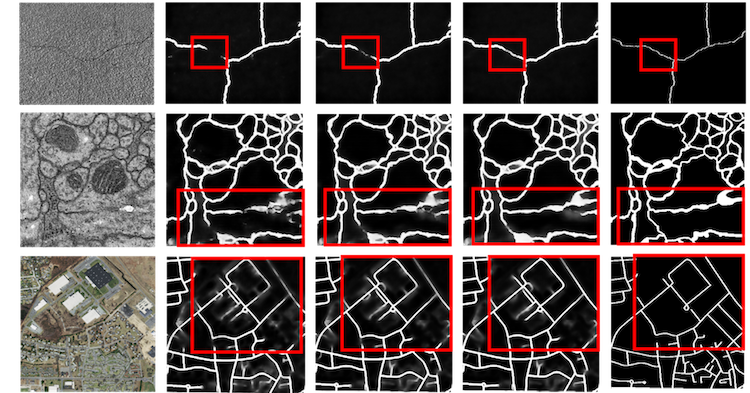}\\
  \end{tabular}
  \vspace{-3mm}
  \caption{\small \textbf{Iterative Refinement.} (a) Prediction after 1, 2 and 3 refinement iterations. The right-most image is the ground-truth. The red boxes highlight parts of the image where refinement is closing gaps.}
    \label{fig:refinement_qualitative}
\end{figure*}


We start by identifying the best-performing configuration for our method. As can be seen in Table~\ref{table:config}(left), using all three first layers  of the VGG network to compute the topology loss yields the best results on the \CRdat{} dataset. It also did so on the other two datasets. Similarly, we evaluated the impact of the number of improvement iterations on the resulting performance on the \EMdat{} dataset, which we present in Table~\ref{table:config}(right). The performance stabilizes after the third iteration. We therefore used three refinement iterations in all further experiments. Note that in Table~\ref{table:config}(right) the first iteration of \OURS{} yields a result that is already better than \OURN{}. This shows that iterative training not only makes it possible to refine the results by iterating at test time, but also yields a better standalone classifier.

We report results of our comparative experiments for the three datasets in Tables~\ref{table:roadsEM}, ~\ref{table:correctness}, and~\ref{table:topology}. 
Even without refinement, our topological loss outperforms all the baselines. Refinement boosts the performance yet further. The differences are greater when using the metrics specifically designed to gauge the quality of linear structures in Table~\ref{table:correctness} and even more when using the topology-based metrics in Table~\ref{table:topology}. This confirms the hypothesis that our contributions improve the quality of the predictions mainly in its topological aspect. The improvement in per-pixel measures, presented in Table~\ref{table:roadsEM} suggests that the improved topology is correlated with better localisation of the predictions.

Finally, we submitted our results to the ISBI challenge server for the \EMdat{} task. We received a foreground-restricted random score of $0.981$. This puts us in first place among algorithms relying on a single classifier without additional processing. In second place is the recent method of~\cite{Shen17}, which achieves the slightly lower score of  $0.978$ even though it relies on a significantly more complex base classifier.

\subsection{Qualitative Results}
\label{sec:qualitative}

Figs.~\ref{fig:cracks_qualitative}, \ref{fig:roads_qualitative}, and~\ref{fig:isbi_qualitative} depict typical results on the three datasets. Note that adding our topology loss term and iteratively refining the delineations makes our predictions more structured and consistently eliminates false positives in the background, without losing the curvilinear structures of interest as shown in Fig.~\ref{fig:refinement_qualitative}. For example, in the aerial images of Fig.~\ref{fig:roads_qualitative}, line-like structures such as roofs and rivers are filtered out because they are not part of the training data, while the roads are not only preserved but also enhanced by closing small gaps. In the case of neuronal membranes, the additional topology term eliminates false positives corresponding to cell-like structures such as mitochondria.


\section{Conclusion}

We have introduced a new loss term that accounts for topology of curvilinear structures by exploiting their higher-level features. We have further improved it by introducing a recursive refinement stage that does not increase the number of parameters to be learned. Our approach is generic and can be used for detection of many types of linear structures including roads and cracks in natural images and neuronal membranes in micrograms. We have relied on the U-Net to demonstrate it but it could be used in conjunction with any other network architecture. In future work, we will explore the use of adversarial networks to adapt our measure of topological similarity and learn more discriminative features.


\begin{thebibliography}{10}\itemsep=-1pt

\bibitem{arganda15}
I.~Arganda-Carreras, S.~Turaga, D.~Berger, D.~CiresŸan, A.~Giusti,
  L.~Gambardella, J.~Schmidhuber, D.~Laptev, S.~Dwivedi, J.~Buhmann, T.~Liu,
  M.~Seyedhosseini, T.~Tasdizen, L.~Kamentsky, R.~Burget, V.~Uher, X.~Tan,
  C.~Sun, T.~Pham, E.~Bas, M.~Uzunbas, A.~Cardona, J.~Schindelin, and S.~Seung.
\newblock {Crowdsourcing the creation of image segmentation algorithms for
  connectomics}.
\newblock {\em Frontiers in Neuroanatomy}, page 142, 2015.

\bibitem{Chai13}
D.~Chai, W.~Forstner, and F.~Lafarge.
\newblock {Recovering Line-networks in Images by Junction-Point Processes}.
\newblock In {\em Conference on Computer Vision and Pattern Recognition}, 2013.

\bibitem{Cicek16}
{\"O}.~{\c C}i{\c c}ek, A.~Abdulkadir., S.~Lienkamp, T.~T.~Brox, and
  O.~Ronneberger.
\newblock {3D U-Net: Learning Dense Volumetric Segmentation from Sparse
  Annotation}.
\newblock {\em arXiv Preprint}, June 2016.

\bibitem{Ganin2014}
Y.~Ganin and V.~S. Lempitsky.
\newblock N4-fields: Neural network nearest neighbor fields for image
  transforms.
\newblock In {\em ACCV}, 2014.

\bibitem{Goodfellow14}
I.~Goodfellow, J.~Pouget-Abadie, M.~Mirza, B.~Xu, D.~Warde-Farley, S.~Ozair,
  A.~Courville, and Y.~Bengio.
\newblock {Generative adversarial nets}.
\newblock In {\em Advances in Neural Information Processing Systems}, pages
  2672--2680, 2014.

\bibitem{Huang09}
X.~Huang and L.~Zhang.
\newblock {Road Centreline Extraction from High-Resolution Imagery Based on
  Multiscale Structural Features and Support Vector Machines}.
\newblock {\em International Journal of Remote Sensing}, 30:1977--1987, 2009.

\bibitem{Isola17}
S.~Isola, J.~Zhu, T.~Zhou, and A.~Efros.
\newblock {Image-to-Image Translation with Conditional Adversarial Networks}.
\newblock In {\em Conference on Computer Vision and Pattern Recognition}, 2017.

\bibitem{Januszewski16}
M.~Januszewski, J.~Maitin-Shepard, P.~P.~Li, J.~Kornfeld, W.~Denk, and V.~Jain.
\newblock {Flood-Filling Networks}.
\newblock {\em arXiv Preprint}, 2016.

\bibitem{Kingma15}
D.~Kingma and J.~Ba.
\newblock {Adam: {A} Method for Stochastic Optimisation}.
\newblock In {\em International Conference for Learning Representations}, 2015.

\bibitem{Law08}
M.~Law and A.~Chung.
\newblock {Three Dimensional Curvilinear Structure Detection Using Optimally
  Oriented Flux}.
\newblock In {\em European Conference on Computer Vision}, 2008.

\bibitem{Maninis16}
K.~Maninis, J.~Pont-Tuset, P.~Arbel\'{a}ez, and L.~V. Gool.
\newblock {Deep Retinal Image Understanding}.
\newblock In {\em Conference on Medical Image Computing and Computer Assisted
  Intervention}, 2016.

\bibitem{Mattyus17}
G.~Mattyusand, W.~L. R., and Urtasun.
\newblock {DeepRoadMapper: Extracting Road Topology From Aerial Images}.
\newblock In {\em International Conference on Computer Vision}, 2017.

\bibitem{Mnih13}
V.~Mnih.
\newblock {\em {Machine Learning for Aerial Image Labeling}}.
\newblock PhD thesis, University of Toronto, 2013.

\bibitem{Mnih10}
V.~Mnih and G.~Hinton.
\newblock {Learning to Detect Roads in High-Resolution Aerial Images}.
\newblock In {\em European Conference on Computer Vision}, 2010.

\bibitem{Mnih12}
V.~Mnih and G.~Hinton.
\newblock {Learning to Label Aerial Images from Noisy Data}.
\newblock In {\em International Conference on Machine Learning}, 2012.

\bibitem{Newell16}
A.~Newell, K.~Yang, and J.~Deng.
\newblock {Stacked Hourglass Networks for Human Pose Estimation}.
\newblock In {\em European Conference on Computer Vision}, 2016.

\bibitem{Pinheiro14}
P.~Pinheiro and R.~Collobert.
\newblock {Recurrent Neural Networks for Scenel Labelling}.
\newblock In {\em International Conference on Machine Learning}, 2014.

\bibitem{Ronneberger15}
O.~Ronneberger, P.~Fischer, and T.~Brox.
\newblock {{U-Net}: Convolutional Networks for Biomedical Image Segmentation}.
\newblock In {\em Conference on Medical Image Computing and Computer Assisted
  Intervention}, 2015.

\bibitem{Russakovsky15}
O.~Russakovsky, J.~Deng, H.~Su, J.~Krause, S.Satheesh, S.~Ma, Z.~Huang,
  A.~Karpathy, A.~Khosla, M.~Bernstein, A.~Berg, and L.~Fei-Fei.
\newblock {Imagenet Large Scale Visual Recognition Challenge}.
\newblock {\em International Journal of Computer Vision}, 115(3):211--252,
  2015.

\bibitem{Seyedhosseini13}
M.~Seyedhosseini, M.~Sajjadi, and T.~Tasdizen.
\newblock {Image Segmentation with Cascaded Hierarchical Models and Logistic
  Disjunctive Normal Networks}.
\newblock In {\em International Conference on Computer Vision}, 2013.

\bibitem{Shen17}
W.~Shen, B.~Wang, Y.~Jiang, Y.~Wang, and A.~L. Yuille.
\newblock {Multi-stage Multi-recursive-input Fully Convolutional Networks for
  Neuronal Boundary Detection}.
\newblock 2017.

\bibitem{Simonyan15}
K.~Simonyan and A.~Zisserman.
\newblock {Very Deep Convolutional Networks for Large-Scale Image Recognition}.
\newblock In {\em International Conference for Learning Representations}, 2015.

\bibitem{Sironi16a}
A.~Sironi, E.~Turetken, V.~Lepetit, and P.~Fua.
\newblock {Multiscale Centerline Detection}.
\newblock {\em IEEE Transactions on Pattern Analysis and Machine Intelligence},
  38(7):1327--1341, 2016.

\bibitem{Stoica04}
R.~Stoica, X.~Descombes, and J.~Zerubia.
\newblock {A Gibbs Point Process for Road Extraction from Remotely Sensed
  Images}.
\newblock {\em International Journal of Computer Vision}, 57(2):121--136, 2004.

\bibitem{Tu09}
Z.~Tu and X.~Bai.
\newblock {Auto-Context and Its Applications to High-Level Vision Tasks and 3D
  Brain Image Segmentation}.
\newblock {\em IEEE Transactions on Pattern Analysis and Machine Intelligence},
  2009.

\bibitem{Turetken13c}
E.~Turetken, C.~Becker, P.~Glowacki, F.~Benmansour, and P.~Fua.
\newblock {Detecting Irregular Curvilinear Structures in Gray Scale and Color
  Imagery Using Multi-Directional Oriented Flux}.
\newblock In {\em International Conference on Computer Vision}, December 2013.

\bibitem{Turetken16a}
E.~Turetken, F.~Benmansour, B.~Andres, P.~Glowacki, H.~Pfister, and P.~Fua.
\newblock {Reconstructing Curvilinear Networks Using Path Classifiers and
  Integer Programming}.
\newblock {\em IEEE Transactions on Pattern Analysis and Machine Intelligence},
  38(12):2515--2530, 2016.

\bibitem{Wegner13}
J.~Wegner, J.~Montoya-Zegarra, and K.~Schindler.
\newblock {A Higher-Order CRF Model for Road Network Extraction}.
\newblock In {\em Conference on Computer Vision and Pattern Recognition}, 2013.

\bibitem{Wiedemann98}
C.~Wiedemann, C.~Heipke, H.~Mayer, and O.~Jamet.
\newblock {Empirical Evaluation Of Automatically Extracted Road Axes}.
\newblock In {\em Empirical Evaluation Techniques in Computer Vision}, pages
  172--187, 1998.

\bibitem{Zhou12}
S.~K. Zhou, C.~Tietjen, G.~Soza, A.~Wimmer, C.~Lu, Z.~Puskas, D.~Liu, and
  D.~Wu.
\newblock {A Learning Based Deformable Template Matching Method for Automatic
  Rib Centerline Extraction and Labeling in CT Images}.
\newblock In {\em Conference on Computer Vision and Pattern Recognition}, 2012.

\bibitem{Zou12}
Q.~Zou, Y.~Cao, Q.~Li, Q.~Mao, and S.~Wang.
\newblock {CrackTree: Automatic crack detection from pavement images.}
\newblock {\em Pattern Recognition Letters}, 33(3):227--238, 2012.

\end{thebibliography}

\end{document}